\documentclass{article}
\pdfoutput=1
\usepackage{microtype}
\usepackage{graphicx}
\usepackage{subfigure}
\usepackage{booktabs} 

\usepackage[utf8]{inputenc}
\usepackage{amsmath}
\usepackage{amssymb}
\usepackage{booktabs}
\usepackage{tikz}
\usepackage{xcolor}
\usepackage{pifont}
\usepackage{multirow}
\usepackage[switch]{lineno}
\usepackage[T1]{fontenc}

\usepackage{hyperref}

\usepackage[accepted]{icml2021}

\icmltitlerunning{Graph Mixture Density Networks}

\newcommand{\GMDN}{\textsc{GMDN}}
\newcommand{\MDN}{\textsc{MDN}}
\newcommand{\DGN}{\textsc{DGN}}
\newcommand{\RAND}{\textsc{RAND}}
\newcommand{\HIST}{\textsc{HIST}}

\definecolor{applegreen}{rgb}{0.55, 0.71, 0.0}
\definecolor{ao(english)}{rgb}{0.0, 0.5, 0.0}
\definecolor{darkpastelred}{rgb}{0.76, 0.23, 0.13}

\newcommand{\ie}{i.e., }
\newcommand{\eg}{e.g., }
\newcommand{\quotes}[1]{``#1''}
\newcommand{\revise}[1]{#1}
\newcommand{\present}{\textcolor{ao(english)}{\ding{51}}}
\newcommand{\absent}{\textcolor{darkpastelred}{\ding{55}}}

\begin{document}

\twocolumn[
\icmltitle{Graph Mixture Density Networks}




\begin{icmlauthorlist}
\icmlauthor{Federico Errica}{pi}
\icmlauthor{Davide Bacciu}{pi}
\icmlauthor{Alessio Micheli}{pi}
\end{icmlauthorlist}

\icmlaffiliation{pi}{Department of Computer Science, University of Pisa}

\icmlcorrespondingauthor{Federico Errica}{federico.errica@phd.unipi.it}
\icmlcorrespondingauthor{Davide Bacciu}{bacciu@di.unipi.it}
\icmlcorrespondingauthor{Alessio Micheli}{micheli@di.unipi.it}

\icmlkeywords{Graph Representation Learning, Probabilistic Models, Neural Networks}

\vskip 0.3in
]



\printAffiliationsAndNotice{}  

\begin{abstract}
We introduce the Graph Mixture Density Networks, a new family of machine learning models that can fit multimodal output distributions conditioned on graphs of arbitrary topology. By combining ideas from mixture models and graph representation learning, we address a broader class of challenging conditional density estimation problems that rely on structured data. In this respect, we evaluate our method on a new benchmark application that leverages random graphs for stochastic epidemic simulations. We show a significant improvement in the likelihood of epidemic outcomes when taking into account both multimodality and structure. The empirical analysis is complemented by two real-world regression tasks showing the effectiveness of our approach in modeling the output prediction uncertainty. Graph Mixture Density Networks open appealing research opportunities in the study of structure-dependent phenomena that exhibit non-trivial conditional output distributions.
\end{abstract}

\section{Introduction}

Approximating the distribution of a target value $y$ conditioned on an input $x$ is at the core of supervised learning tasks. When trained using common losses such as Mean Square Error for regression or Cross-Entropy for classification, supervised methods are known to approximate the expected conditional distribution of the target given the input, that is, $\langle y|x \rangle$ \cite{bishop_mixture_1994}. This is standard practice when the target distribution is unimodal and slight variations in the target value are mostly due to random noise.

Still, when the target distribution of a regression problem is not unimodal, most machine learning methods fail to represent it correctly by predicting an averaged value. As a matter of fact, a multimodal target distribution associates more than one likely outcome with a given input sample\revise{, and in this case one usually talks about solving a conditional density estimation problem.} To address this, the Mixture Density Network (MDN) \cite{bishop_mixture_1994} was proposed to approximate arbitrarily complex conditional target distributions, and it finds application in robotics \cite{choi_uncertainty_2018}, epidemiology \cite{davis_use_2020} and finance \cite{schittenkopf_volatility_1998}, to name a few. \MDN{}s were designed for input data of vectorial nature, but often real-world problems deal with relational data where the structure substantially impacts the possible outcomes. For instance, this is especially true in epidemiology \cite{opuszko_impact_2013}.

For more than twenty years, researchers have put great effort into the adaptive processing of graphs (see recent surveys of \citet{bacciu_gentle_2020,wu_comprehensive_2020}). The goal is to infer the best representation of a structured sample for a given task via different neighborhood aggregation schemes, graph coarsening, and information propagation strategies. It is easy to find applications that benefit from the adaptive processing of structured data, such as drug design \cite{podda_deep_2020}, classification in social networks \cite{yang_revisiting_2016}, and natural language processing \cite{beck_graph--sequence_2018}.

Our main contribution is the proposal of a hybrid approach to handle multimodal target distributions within machine learning methods for graphs, called Graph Mixture Density Network (\GMDN{}). This model outputs a multimodal distribution, conditioned on an input graph, for either the whole structure or its entities. \revise{For instance, given an observable input graph $x$, \GMDN{} is trained to approximate the (possibly multimodal) distribution associated with the target random variable $y$ via maximum likelihood estimation. The likelihood is the usual metric to be optimized for density estimation tasks \cite{nowicki_estimation_2001}, and it tells us how well the model is fitting the empirical data distribution. Recall that, in general, it does not suffice to predict a single output value like in \quotes{standard} regression problems \cite{bishop_mixture_1994} to solve this kind of tasks; for this reason, \GMDN{} extends the capabilities of deep learning models for graphs whose output is restricted to unimodal distributions.}

We test GMDN on a novel benchmark application introduced in this paper, comprising large epidemiological simulations\footnote{\url{https://github.com/diningphil/graph-mixture-density-networks}} where both structure and multimodality play an essential role in determining the outcome of an epidemic. Results show that \GMDN{} produces a significantly improved likelihood. Then, we evaluate our model on two real-world chemical graph regression tasks to show how \GMDN{} can better model the uncertainty in the output prediction, \ie the model reveals that there might be more than one admissible chemical property value associated with a given input molecule representation.

\section{Related Works}
\label{sec:related-works}
The problem of training a network to output a conditional multimodal distribution, \ie a distribution with one or more modes, has been studied for 30 years. The Mixture of Experts (MoE) model \cite{jacobs_adaptive_1991, jordan_hierarchical_1994} is one of the first proposals that can achieve the goal, even though it was originally meant for a different purpose. The MoE consists of a multitude of neural networks, also called local experts, each being expected to solve a specific sub-task. In addition, an MoE uses a gating network to weigh the local experts' contributions for each input. This way, the model selects the experts that are most likely to make the correct prediction. The overall MoE output is then the weighted combination of the local experts' outputs; the reader is referred to \citet{yuksel_twenty_2012} and \citet{masoudnia_mixture_2014} for comprehensive surveys on this topic. Lastly, notice that the MoE imposes soft competition between the experts, but that may not be necessary when modeling the conditional distribution of the data. 

The Mixture Density Network (MDN) of \citet{bishop_mixture_1994}, instead, reduces the computational burden of training an MoE while allowing the different experts, now called sub-networks, to cooperate. An MDN is similar to an MoE model, but it has subtle differences. First, the input is transformed into a hidden representation that is shared between simpler sub-networks, thus increasing the overall efficiency. Secondly, this representation is used to produce the gating weights as well as the parameters of the different output distributions. Hence, the initial transformation should encode all the information needed to solve the task into said representation. As the computational costs of processing the input grow, so does an MDN's efficiency compared to an MoE. This is even more critical when the input is structured, such as a sequence or a graph, as it requires more resources to be processed.

In terms of applications, \MDN{}s have been recently applied to epidemic simulation prediction \cite{davis_use_2020}. The goal is to predict the multimodal distribution of the total number of infected cases under a compartmental model such as the stochastic Susceptible-Infectious-Recovered (SIR) model \cite{kermack_contribution_1927}. In the paper, the authors show that, given samples of SIR simulations with different infectivity and recovery parameters, the MDN could approximate the conditioned output distribution using a mixture of binomials. This result is a remarkable step in approximating way more complex compartmental models in a fraction of the time originally required, similarly to what has been done, for example, in material sciences \cite{pilania_accelerating_2013}. However, the work of \citet{davis_use_2020} makes the strong assumption that the infected network is a complete graph. In fact, as stated in \cite{opuszko_impact_2013}, arbitrary social interactions in the network play a fundamental role in the spreading of a disease. As such, predictive models should be able to take them into account. 

The automatic and adaptive extraction of relational information from graph-structured data is another long-standing research topic \cite{sperduti_supervised_1997,frasconi_general_1998,micheli_neural_2009,scarselli_graph_2009} that has found widespread application in social sciences, chemistry, and bioinformatics. In the recent past, graph kernels \cite{ralaivola_graph_2005,vishwanathan_graph_2010} were the main methodology to process structural information; while still effective and powerful, the drawback of graph kernels is the computational costs required to compute similarity scores between pairs of graphs. Nowadays, the ability to efficiently process graphs of arbitrary topology is made possible by a family of models called Deep Graph Networks\footnote{\citet{bacciu_gentle_2020} introduced the term in lieu of GNN to avoid ambiguities and to consider deep non-neural models as well.} (\DGN{}s). A DGN stacks graph convolutional layers, which aggregate each node's neighboring states, to propagate information across the graph. The number of layers reflects the amount of contextual information that propagates \cite{micheli_neural_2009}, very much alike to receptive fields of convolutional neural networks \cite{lecun_convolutional_1995}. There is an increasingly growing literature on the topic which is not covered in this work, so we refer the reader to recent introductory texts and surveys \cite{bronstein_geometric_2017,battaglia_relational_2018,bacciu_gentle_2020,wu_comprehensive_2020}.

For the above reasons, we propose the Graph Mixture Density Networks to combine the benefits of \MDN{}s and \DGN{}s. To the best of our knowledge, this is the first \DGN{} that can learn multimodal output distributions conditioned on arbitrary input graphs.

\begin{figure*}[t!]
    \vskip 0.1in
    \centering
    \includegraphics[width=0.9\textwidth]{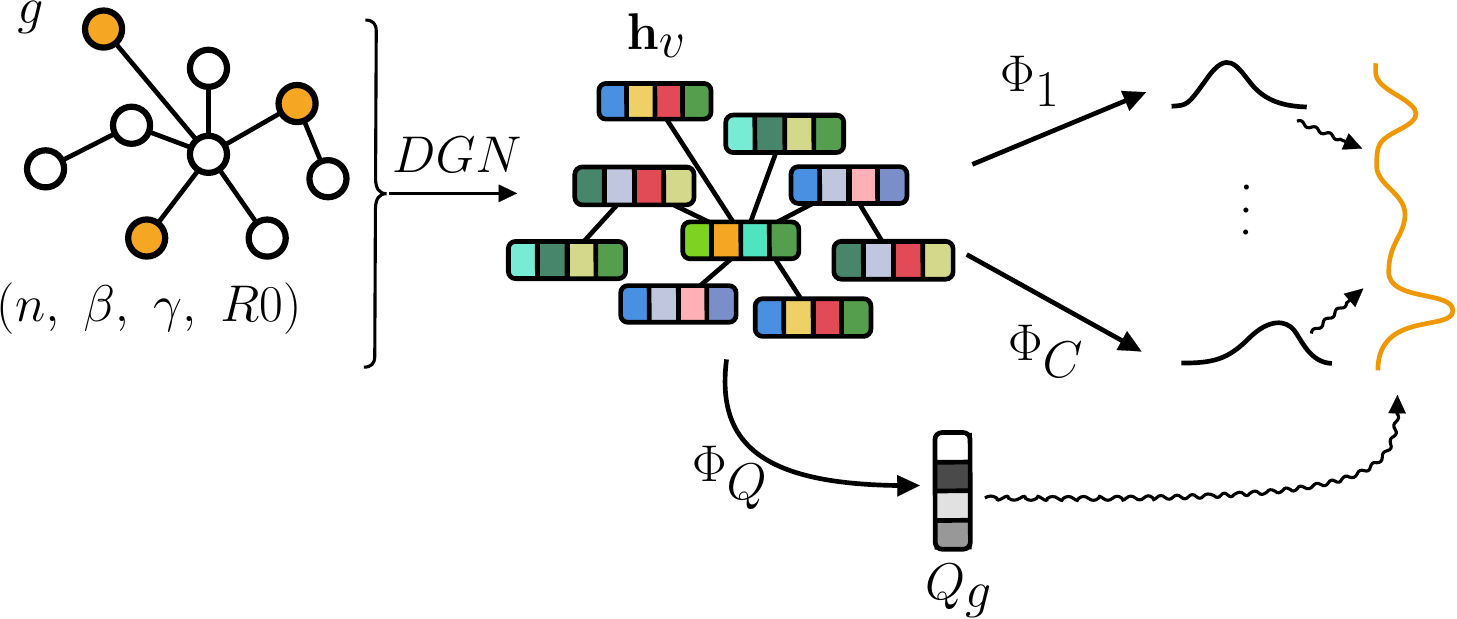}
    \caption{From a high-level perspective, a \DGN{} transforms each node $v$ of the input graph $g$ into a hidden representation $\mathbf{h}_v$ that encodes the structural information surrounding that node. Then, in this work, a subsequent transformation $\Phi_Q$ generates the mixing probability vector $Q_g \in [0,1]^C$ that combines the $C$ different distributions produced by the sub-networks $\Phi_1,\dots,\Phi_C$. Similarly to \MDN{}s, the input's first transformation is shared between the sub-networks. For example, suppose we were to predict the outcome of a stochastic SIR simulation. In that case, orange round nodes might represent initially infected entities in a network of size $n$, and $\beta, \gamma, R0$ would be simulation-specific node attributes.}
    \label{fig:high-level}
    \vskip 0.1in
\end{figure*}
\section{Graph Mixture Density Networks}
\label{sec:model}

A graph is defined as a tuple $g = (\mathcal{V}_g,\mathcal{E}_g,\mathcal{X}_g)$ where $\mathcal{V}_g$ is the set of \emph{nodes} representing entities, $\mathcal{E}_g$ is the set of \emph{edges} that connect pairs of nodes, and $\mathcal{X}_g$ denotes the (optional) node attributes. For the purpose of this work, we do not use edge attributes even though the approach can be straightforwardly extended to consider them.

The task under consideration is a supervised conditional density estimation (CDE) problem. We aim to learn the conditional distribution $P(y_g|g)$, with $y_g$ being the continuous target label(s) associated with an input graph $g$ in the dataset $\mathcal{D}$. We assume the target distribution to be multimodal, and as such it cannot be well modeled by current DGNs due to the aforementioned averaging effects. Therefore, we borrow ideas from the Mixture Density Network \cite{bishop_mixture_1994} and extend the family of deep graph networks with multimodal output capabilities.

From a high-level perspective, we seek a DGN that performs an isomorphic transduction \cite{frasconi_general_1998} to obtain node representations $\mathbf{h}_{\mathcal{V}_g}=\{\mathbf{h}_v \in \mathbb{R}^d, d \in \mathbb{N}, \ \forall v \in g\}$ as well as a set of \quotes{mixing weights} $Q_g \in [0,1]^C$ that sum to 1, where $C$ is the number of unimodal output distributions we want to mix. Given $\mathbf{h}_{\mathcal{V}_g}$, we then apply $C$ different sub-networks $\Phi_1,\dots,\Phi_C$ that produce the parameters $\theta_1,\dots,\theta_C$ of $C$ output distributions, respectively.

In principle, we can mix distributions from different families, \revise{but this poses several issues, such as their choice and how many of them to use for each family. In light of this,} we stick to a single family for simplicity of exposition. Finally, combining the $C$ unimodal output distributions with the mixing weights $Q_g$ produces a multimodal output distribution. We sketch the overall process in Figure \ref{fig:high-level} for the specific case of epidemic simulations.

\begin{figure}[t]
    \vskip 0.1in
    \centering
    \includegraphics[width=0.5
    \columnwidth]{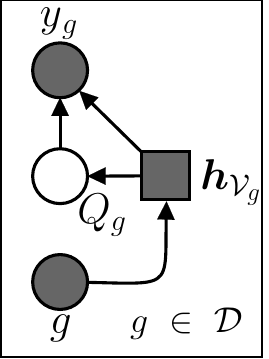}
    \caption{The model can be graphically represented as a Bayesian network where round white (dark) nodes are unobserved (observed) random variables. For each graph $g$ in the dataset $\mathcal{D}$, we introduce the latent variable $Q_g$ via marginalization. This allows us to break the computation of  $P(y_g|g)$ in two steps. The first step encodes the graph information into deterministic node states $\mathbf{h}_{\mathcal{V}_g}$ and produces the posterior distribution $P(Q_g|g)$. In the second and final step, we output the emission distributions $P(y_g|Q_g=i, g), i=1,\dots,C$. The result is a mixture model conditioned on the input structure.}
    \label{fig:graphical-model}
    \vskip 0.1in
\end{figure}

More formally, we learn the conditional distribution $P(y_g|g)$ using the Bayesian network of Figure \ref{fig:graphical-model}. Here, round white (dark) nodes represent unobserved (observed) random variables, and larger squares indicate deterministic outputs. The mixing weights $Q_g$ are modeled as a categorical distribution with $C$ possible states. 

We solve the CDE problem by maximum likelihood estimation (MLE). The likelihood, \ie $P(y|g)$, is the usual quantity to be maximized. It reflects the probability that an output $y$ is generated from a graph $g$. Given an hypotheses space $\mathcal{H}$, we seek the MLE hypothesis:
\begin{align}
& h_{MLE} =  arg\max_{h\in\mathcal{H}}P(\mathcal{D}|h) = \nonumber \\
& = arg\max_{h\in\mathcal{H}} \prod_{g\in\mathcal{D}}\sum_{i=1}^C P(y_g|Q^i_g, g)P(Q^i_g|g),
\label{eq:map-equation}
\end{align}
where we introduced the latent variable $Q_g$ via marginalization whose \textit{i}-th component is $Q_g^i$. In particular, we will model the distributions of Equation \ref{eq:map-equation} by means of deep graph networks, which allow great flexibility with respect to the input structure \revise{and invariance to graph automorphism. This way, we are able to approximate probabilities that are conditioned on a variable number of graph nodes and edges.}

As mentioned earlier, a deep graph network encodes the input graph into node representations $\mathbf{h}_{\mathcal{V}_g}$. Generally speaking, this encoder stacks multiple layers of graph convolutions to generate intermediate node states $\mathbf{h}_v^{\ell}$ at each layer $\ell = 1,\dots,L$:
\begin{align}
\mathbf{h}_v^{\ell+1} = \phi^{\ell+1} \Big(\mathbf{h}_v^\ell,\ \Psi(\{\psi^{\ell+1}(\mathbf{h}_u^{\ell}) \mid u \in \mathcal{N}_v\} ) \Big),
\label{eq:aggregation}
\end{align}
where $\phi$ and $\psi$ are (possibly non-linear) functions, and $\Psi$ is a permutation invariant function applied to node $v$'s neighborhood $\mathcal{N}_v$. Usually, the final node representation $\mathbf{h}_v$ is given by $\mathbf{h}^L_v$ or, alternatively, by the concatenation of all intermediate states. The convolution of the Graph Isomorphism Network (GIN) \citep{xu_how_2019} is a particular instance of Equation \ref{eq:aggregation} that we will use in our experiments to compute graph-related probabilities, as these need to be permutation invariant with respect to the node ordering.

In graph-prediction, representations $\boldsymbol{h}_{\mathcal{V}_g}$ have to be further aggregated with another permutation invariant function $\Psi_g$
\begin{align}
    \mathbf{h}_g = r_g(\mathbf{h}_{\mathcal{V}_g})=\Psi_g \Big( \{ f_r(\mathbf{h}_v) \mid v \in \mathcal{V}_g \} \Big),
\label{eq:readout}
\end{align}
where $f_r$ could be a linear model or a Multi-Layer Perceptron. Equation \ref{eq:readout} is often referred to as the \quotes{readout} phase. Instead, the mixing weights can be computed using a readout $r^Q_g$ as follows: 
\begin{align}
 P(Q_g|g) = \sigma(r^Q_g(\mathbf{h}_{\mathcal{V}_g})),
\label{eq:mixing-weights}
\end{align}
where $\sigma$ is the softmax function over the components of the aggregated vector.

To learn the emission $P(y_g|Q^i_g,g), \ i=1,\dots,C$, we have to implement a sub-network $\Phi_i$ that outputs the parameters of the chosen distribution. For instance, if the distribution is a multivariate Gaussian we have
\begin{align}
\boldsymbol{\mu_i},\boldsymbol{\Sigma_i} = \Phi_i(\mathbf{h}_g) = f_i(r^i_g(\mathbf{h}_{\mathcal{V}_g})),
\label{eq:output-parameters}
\end{align}with $f_i$ being defined as $f_r$ above. Note that node-prediction tasks do not need a global readout phase, so Equations \ref{eq:mixing-weights} and \ref{eq:output-parameters} are directly applied to $\mathbf{h}_v \ \forall v \in \mathcal{V}_g$.

Differently from the Mixture of Experts, which would require a new DGN encoder for each output distribution $i$, we follow the Mixture Density Network approach and share $\mathbf{h}_{\mathcal{V}_g}$ between the sub-networks. This form of weight sharing reduces the number of parameters and pushes the model to extract all the relevant structural information into $\mathbf{h}_{\mathcal{V}_g}$. Furthermore, using multiple DGN encoders can become computationally intractable for large datasets.

\paragraph{Training.} We train the Graph Mixture Density Network using the Expectation-Maximization (EM) framework \cite{dempster_maximum_1977} for MLE estimation. We choose EM for the local convergence guaranteees that it offers with respect to other optimizers, and since its effectiveness has already been proved on probabilistic graph models \cite{bacciu_contextual_2018, bacciu_probabilistic_2020}. Indeed, by introducing the usual indicator variable $z^g_i \in \mathcal{Z}$, which is one when graph $g$ is in latent state $i$, we can compute the lower bound of the log-likelihood as in standard mixture models \cite{jordan_hierarchical_1994,corduneanu_variational_2001}:
\begin{align}
    & \mathbb{E}_{\mathcal{Z}|\mathcal{D}}[\log \mathcal{L}_c(h|\mathcal{D})] = \nonumber \\
    & = \sum_{g \in \mathcal{D}}\sum_{i=1}^C E[z_i^g\vert\mathcal{D}] \log \Big(P(y_g|Q^i_g,g)P(Q_g|g) \Big)
    \label{eq:e-step}
\end{align}
where $\log \mathcal{L}_c(h|\mathcal{D})$ is the complete log likelihood.

The E-step of the EM algorithm can be performed analytically by computing the posterior probability of the indicator variables:
\begin{align}
    E[z_i^g\vert\mathcal{D}]=P(z^g_i=1|g)=\frac{1}{Z}P(y_g|Q^i_g,g)P(Q_g|g) 
\end{align}
where $Z$ is the usual normalization term obtained via straightforward marginalization. On the other hand, we do not have closed-form solutions for the M-step because of the non-linear functions used. Hence, we perform the M-step using gradient ascent to maximize Equation \ref{eq:e-step}. The resulting algorithm is known as Generalized EM (GEM) \cite{dempster_maximum_1977}. GEM still guarantees convergence to a local minimum if each optimization step improves Equation \ref{eq:e-step}. Finally, we introduce an optional Dirichlet regularizer $\pi$ with hyper-parameter $\boldsymbol{\alpha}=(\alpha_1,\dots,\alpha_C)$ on the distribution $P(Q_g|g)$. The prior distribution serves to prevent the posterior probability mass of the from collapsing onto a single state. This is a well-known problem that has been addressed in the literature through specific constraints \cite{eigen_learning_2013} or entropic regularization terms \cite{pereyra_regularizing_2017}. Here, the objective to be maximized becomes
\begin{align}
\underbrace{\mathbb{E}_{\mathcal{Z}|\mathcal{D}}[\log \mathcal{L}_c(h|\mathcal{D})]}_\text{original objective} +  \underbrace{\sum_{g \in \mathcal{D}}\log \pi(Q_g|\boldsymbol{\alpha})}_\text{Dirichlet regularizer},
\label{eq:objective}
\end{align}
where we note that $\boldsymbol{\alpha}=\boldsymbol{1}^C$ corresponds to a uniform prior, \ie no regularization. To conclude, maximizing Equation \ref{eq:objective} still preserves the convergence guarantees of GEM if the original objective increases at each step.

\section{Experiments}
\label{sec:experiments}

This section thoroughly describes the datasets, experiments, evaluation process and hyper-parameters used. This work aims at showing that \GMDN{} can fit multimodal distributions conditioned on a graph better than using \MDN{}s or \DGN{}s individually. To do so, we publicly release large datasets of stochastic SIR simulations whose results depend on the underlying network, rather than assuming uniformly distributed connections as in \citet{davis_use_2020}. We generate random graphs using the Barabasi-Albert (BA) \cite{barabasi_emergence_1999} and Erdos-Renyi (ER) \cite{bollobas_random_2001} models. While ER graphs do not preserve social networks' properties, here we are interested in the emergence of multimodal outcome distributions rather than biological plausibility. \revise{That said, future investigation will cover more realistic cases, for instance using the Block Two-Level Erdos–Renyi model \cite{seshadhri_community_2012}}. We expect \GMDN{} to perform better because it takes both multimodality and structure into account during training. Moreover, we analyze whether training on a particular family of graphs exhibits transfer properties; if that is the case, then the model has learned how to make informed predictions about different (let alone completely new) structures. At last, we apply the model on two molecular graph regression benchmarks to analyze the performances of GMDN on real-world data.

\begin{figure}[t]
\vskip 0.1in
\centering
\includegraphics[width=\columnwidth]{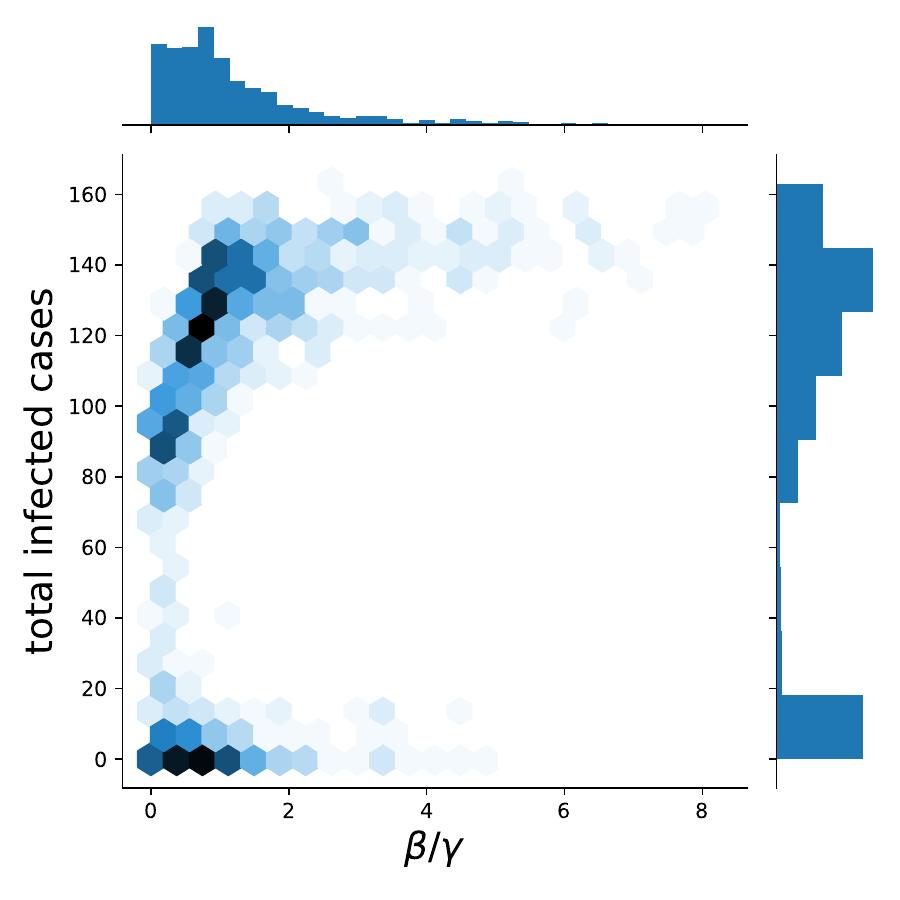}
\caption{Given a single network and specific choices for $R0=\beta/\gamma$, the repeated simulation of the stochastic SIR model is known to produce different outcomes. Here we plot the outcome distributions of 1000 SIR simulations on an Erdos-Renyi network of size 200. \revise{We follow \citet{davis_use_2020} and sample $\beta$ and $\gamma$ uniformly, rather than their ratio, because higher ratios correspond to less interesting behaviors, \ie the distribution becomes unimodal.} Depending on the input structure, the distribution of the total infected cases may be multimodal or not, and the \GMDN{} should recognize this phenomenon. In our simulations, larger networks exhibited less multimodality; hence, without loss of generality, we focus on larger datasets of smaller graphs.}
\label{fig:distribution-infected}
\vskip 0.1in
\end{figure}

\paragraph{Datasets.} We simulated the well-known stochastic SIR epidemiological model on Barabasi-Albert graphs of size 100 (BA-100), generating 100 random graphs for different connectivity values (2, 5, 10 and 20). Borrowing ideas from \citet{davis_use_2020}, for each configuration, we run 100 simulations for each different initial infection probability (1\%, 5\%, 10\%) sampling the infectivity parameter $\beta$ from $[0, 1]$ and the recovery parameter $\gamma$ from $[0.1, 1]$. We also carry out simulations for Erdos-Renyi graphs (ER-100), this time with connectivity parameters 0.01, 0.05, 0.1, and 0.2. The resulting total number of simulations (\ie samples) in each dataset is 120.000, and the goal is to predict the distribution of the total infected cases at the end of a simulation. Node features consist of $\beta$, $\gamma$, their ratio $R0=\beta/\gamma$, a constant value $1$, and a binary value that indicates whether that node is infected or not at the beginning of the simulation. Moreover, to test the transfer learning capabilities of GMDN on graphs with different structural properties (according to the chosen random graph model), we constructed six additional simulation datasets where graphs have different sizes, \ie from 50 to 500. An example of simulation results is summarized in Figure \ref{fig:distribution-infected}; we observe that the outcome distribution of repeated simulations on a single graph leads to a multimodal distribution, in accord with \cite{opuszko_impact_2013}. Therefore, in principle, being able to accurately and efficiently predict the outcome distribution of a (possibly complex) epidemiological model can significantly impact the preparations for an incumbent sanitary emergency.

When dealing with graph regression tasks, especially in the chemical domain, we usually do not expect such a conspicuous emergence of multimodality in the output distribution. Indeed, the properties of each molecule are assumed to be regulated by natural laws, but the information we possess about the input representation may be incomplete and/or noisy. Similarly, the way the model processes the input has an impact on the overall uncertainty; for instance, disregarding bond information makes graphs appear isomorphic to the model while they are indeed not so. As such, knowing the confidence of a trained regressor for a specific outcome becomes invaluable to better understand the data, the model behavior, and, ultimately, to determine the trust we place in each prediction. Therefore, we will evaluate our model on the large chemical benchmarks alchemy\_full \cite{chen_alchemy_2019} and ZINC\_full \cite{irwin_zinc_2012,bresson_two_2019} made of 202579 and 249456 molecules, respectively. The task of both datasets is the prediction of continuous chemical properties (12 for the former and 1 for the latter) associated with each molecule representation (9 and 28 node features, respectively). As in \citet{chen_alchemy_2019}, the GIN convolution used only considers the existence of a bond between atoms. In the considered datasets, this gives rise to isomorphic representations of different molecules when bond types or 3D coordinates are not considered (or ignored by the model). The same phenomena, in different contexts and forms, can occur whenever the original data or its choice of representation lack part of the information to solve a task.

\paragraph{Evaluation Setup.} We assess the performance of different models using a holdout strategy for all datasets ($80\%/10\%/10\%$ split). \revise{Given the size of the datasets, we believe that a simple holdout is sufficient to assess the performances of the different models considered.} To make the evaluation \revise{even} more robust for the epidemic datasets, different simulations about the same graph cannot appear in both training and test splits. The metric of interest is the log-likelihood of the data ($\log \mathcal{L}$), which captures how well we can fit the target distribution and the model's uncertainty with respect to a particular output value. We also report the Mean Average Error (MAE) on the real-world benchmarks for completeness. However, the MAE does not reflect the model's uncertainty about the output, as we will show.

We perform model selection via grid search for all the models presented. \revise{For each of them, we} select the best configuration on the validation set using early stopping with patience \cite{prechelt_early_1998}. \revise{Then}, to avoid an unlucky random initialization of the chosen configuration, we average the model's performance on the unseen test set over ten final training runs. \revise{Similarly to the model selection phase, in these final training runs we use early stopping on a validation set extracted from the training set (10\% of the training data)}.

\paragraph{Baselines and hyper-parameters.}
We compare \GMDN{} against four different baselines. First, \RAND{} predicts the uniform probability over the finite set of possible outcomes, thus providing the threshold log-likelihood score above which predictions are useful. Instead, \HIST{} computes the normalized frequency histogram of the target values given the training data, which is then converted into a discrete probability. While on epidemic simulations we can use the graph's size as the number of histogram bins to use, on the chemical benchmarks this number must be treated as a hyper-parameter and manually cross-validated against the validation set. \HIST{} is used to test whether multimodality is useful when a model does not take the structure into account. Finally, we have \MDN{} and \DGN{}, which are, in a sense, ablated versions of GMDN. Indeed, \MDN{} ignores the input structure, whereas \DGN{} cannot model multimodality. Neural models are trained to output unimodal (\DGN{}) or multimodal (\MDN{}, \GMDN{}) binomial distributions for the epidemic simulation datasets and isotropic Gaussians for the chemical ones. The sub-networks $\Phi_i$ are linear models, and the graph convolutional layer is adapted from \citet{xu_how_2019}.
We conclude the section by listing the hyper-parameters tried for each model:
\begin{itemize}
\setlength\itemsep{-0.4em}
\item \MDN{}: $C$ $\in$ \{2,3,5\}, hidden units per convolution $\in$ \{64\}, neighborhood aggregation $\in$ \{sum\}, graph readout $\in$ \{sum, mean\}, $\boldsymbol{\alpha}$ $\in$ \{$\boldsymbol{1}^C$, $\boldsymbol{1.05}^C$\}, epochs $\in$ \{2500\}, $\Phi_i$ $\in$ \{\text{Linear model}\},  Adam Optimizer with learning rate $\in$ \{0.0001\}, full batch, patience $\in$ \{30\}.
\item \GMDN{}: $C$ $\in$ \{3,5\}, graph convolutional layers $\in$ \{2,5,7\}, hidden units per convolution $\in$ \{64\}, neighborhood aggregation $\in$ \{sum\}, graph readout  $\in$ \{sum, mean\}, $\boldsymbol{\alpha}$ $\in$ \{$\boldsymbol{1}^C$, $\boldsymbol{1.05}^C$\}, epochs $\in$ \{2500\}, $\Phi_i$ $\in$ \{\text{Linear model}\},  Adam Optimizer with learning rate $\in$ \{0.0001\}, full batch, patience $\in$ \{30\}.
\item \DGN{}:  same as \GMDN{} but $C$ $\in$ \{$1$\} (that is, it outputs a unimodal distribution).
\end{itemize}
\revise{Note that we kept the maximum number of epochs intentionally high as we use early stopping to halt training.}\revise{Also, the results of the experiments hold regardless of the \DGN{} variant used, given the fact that \DGN{}s output a single value rather than a complex distribution. In other words, we compare \textit{families} of models rather than specific architectures.}

\section{Results and Discussion}
\label{sec:results}

\begin{table}[t]
\centering
\small
\begin{tabular}{lcccc}
\toprule
\textbf{Model}      & \textbf{\textsc{BA-100}} & \textbf{\textsc{ER-100}} & \textbf{Structure}       & \textbf{Multimodal}        \\ \midrule
\RAND{} & -4.60       & -4.60       &    \absent      &    \absent        \\
\HIST{} & -1.16       & -2.32       &    \absent      &    \present        \\ 
\MDN{}  & -1.17(.05)       & -2.54(.07)       &    \absent     &    \present       \\
\DGN{}  & -0.90(.35)       & -1.96(.16)            &    \present      &    \absent        \\ \midrule
\GMDN{} & \textbf{-0.67}(.02) & \textbf{-1.56}(.04) & \present   &    \present      \\ \bottomrule
\end{tabular}
\caption{Results on BA-100 e ER-100 (12.000 test samples each). A higher log-likelihood corresponds to better performances. \GMDN{} improves the performance on both tasks, showing the advantages of that taking into account both multimodality and structure. Neural models' results are averaged over 10 runs, and standard deviation is reported in brackets.}
\label{tab:results}
\end{table}

\begin{figure*}[t]
\vskip 0.1in
\centering
\begin{subfigure}
  \centering
  \includegraphics[width=0.49\textwidth]{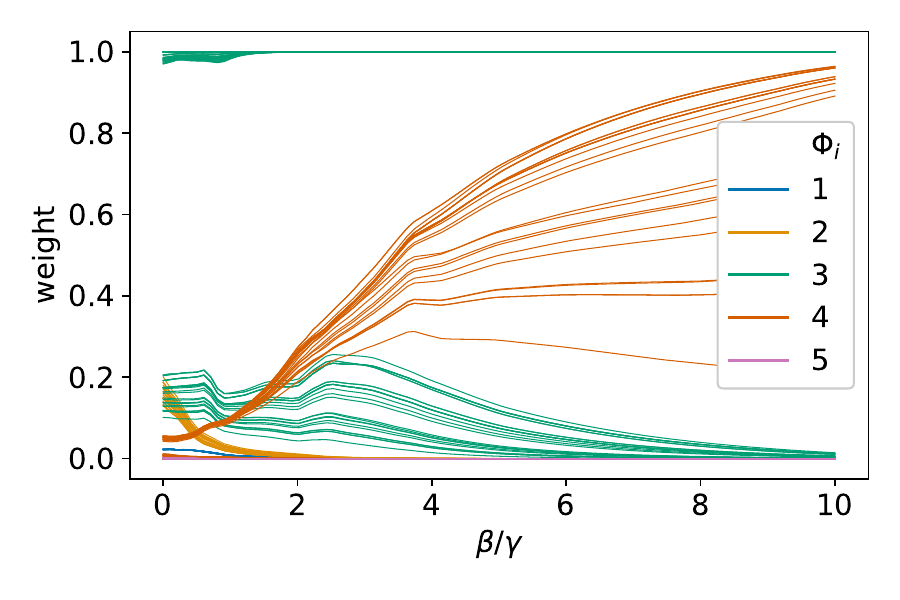}
\end{subfigure}%
\begin{subfigure}
  \centering
  \includegraphics[width=0.49\textwidth]{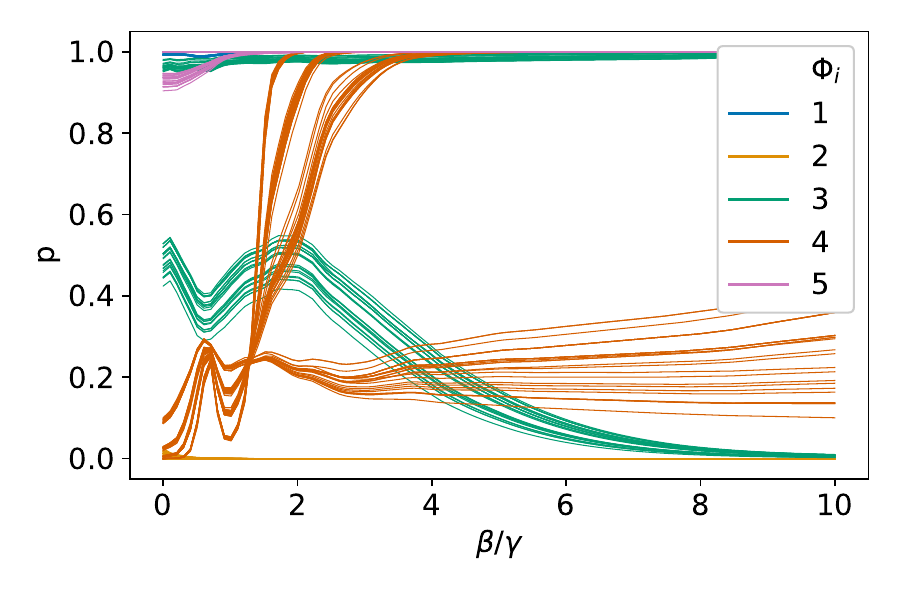}
\end{subfigure}
\caption{The trend of the mixing weights (left) and binomial coefficient (right) for each one of five sub-networks is shown on 100 ER-100 graphs. We vary the ratio between infection and recovery rate to inspect the behavior of the \GMDN{}. Here, we see that sub-network 4 can greatly change the binomial output distribution in a way that depends on the input graph.}
\label{fig:parameters-distributions}
\vskip 0.1in
\end{figure*}

This section discusses our experimental findings. We start from the main empirical study on epidemic simulations, which include CDE results and transferability of the learned knowledge. Then, we report results obtained on the real-world chemical tasks, highlighting the importance of capturing a model's uncertainty about the output predictions.

\paragraph{Epidemic Simulation Results} 
We begin by analyzing the results obtained on BA-100 and ER-100 in Table \ref{tab:results}. We notice that \GMDN{} has better test log-likelihoods than the other baselines, with larger performance gains on ER-100.  Being \GMDN{} the only model that considers both structure and multimodality, such an improvement was expected. However, it is particularly interesting that \HIST{} has a better log-likelihood than \MDN{} on both tasks. By combining this fact with the results of \DGN{}, we come to two conclusions. First, the structural information seems to be the primary factor of performance improvement; this should not come as a surprise since the way an epidemic develops depends on how the network is organized (despite we are not aiming for biological plausibility). Secondly, none of the baselines can get close enough to \GMDN{} on ER-100, indicating that this task is harder to solve by looking individually at structure or multimodality. In this sense, BA-100 might be considered an easier task than ER-100, and this is plausible because emergence of multimodality on the former task seems slightly less pronounced in the SIR simulations. For completeness, we also tested an intermediate baseline where DGN is trained with L1 loss followed by MDN on the graph embeddings. Results displayed a $\log\mathcal{L} \approx -16$ on both datasets, probably because the DGN creates similar graph embeddings for different distributions with the same mean, with consequent severe loss of information.

Similarly to what has been done in \citet{bishop_mixture_1994} and \citet{davis_use_2020}, we analyze how the mixing weights and the distribution parameters vary on a particular \GMDN{} instance. We use $C$=5 and track the behavior of each sub-network for 100 different ER-100 graphs. Figure \ref{fig:parameters-distributions} shows the trend of the mixing weights (left) and of the binomial parameters $p$ (right) for different values of the ratio $R0=\beta/\gamma$. We immediately see that many of the sub-networks are \quotes{shut down} as the ratio grows. In particular, sub-networks 3 and 4 are the ones that control \GMDN{}'s output distribution the most, though for high values of R0 only one sub-network suffices. These observations are concordant with the behavior of Figure \ref{fig:distribution-infected}: when the infectivity rate is much higher than the recovery rate, the target distribution becomes unimodal. The analysis of the binomial parameter for sub-network 4 provides another interesting insight. We notice that, depending on the input graph, the sub-network leads to two possible outcomes: the outbreak of the disease or a partial infection of the network. Note that this is a behavior that \GMDN{} can model whereas the classical \MDN{} cannot.

\begin{figure}[t!]
    \vskip 0.1in
    \centering
    \includegraphics[width=\columnwidth]{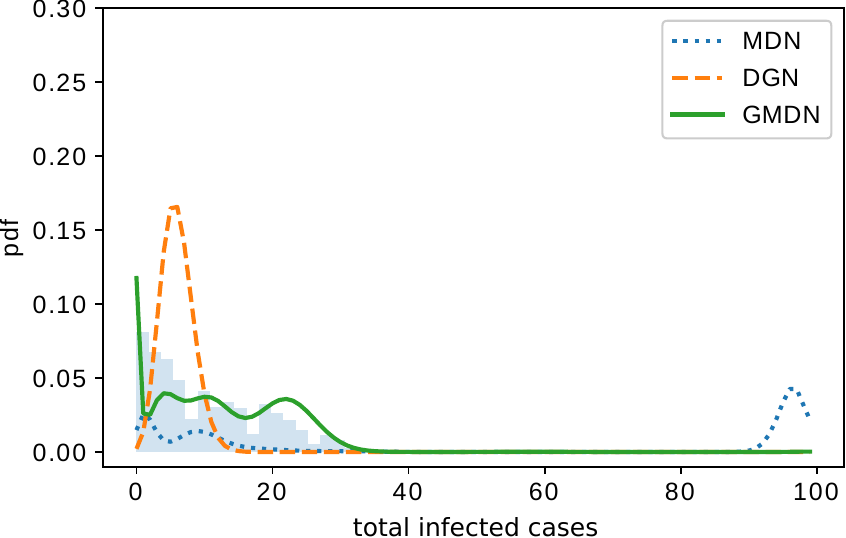}
    \caption{Output distributions of \MDN{}, \DGN{}, and \GMDN{} on an ER graph of size 100. As we can see, the \GMDN{} can provide a rich multimodal distribution conditioned on the structure close to that generated by SIR simulations (blue histogram).}
    \label{fig:erdos-renyi}
    \vskip 0.1in
\end{figure}

To provide further evidence about the benefits of the proposed model, Figure \ref{fig:erdos-renyi} shows the output distributions of \MDN{}, \DGN{} and \GMDN{} for a given sample of the ER-100 dataset. We also plot the result of SIR simulations on that sample as a blue histogram (ground truth). Some observations can be made. First, the \MDN{} places the output probability mass at both sides of the plot. This choice is understandable considering the lack of knowledge about the underlying structure (see also Table \ref{tab:results}) and the fact that likely output values tend to be polarized at the extremes (see \eg Figure \ref{fig:distribution-infected}). Secondly, the \DGN{} can process the structure but cannot model more than one outcome. Therefore, and coherently with \citet{bishop_mixture_1994} for vectorial data, the \DGN{} unique mode lies in between those of \GMDN{} that account for the majority of \GMDN{} probability mass.  In contrast, \GMDN{} produces a multimodal and structure-aware distribution that closely follows the ground truth.

\begin{figure}[t]
    \vskip 0.1in
    \centering
    \includegraphics[width=\columnwidth]{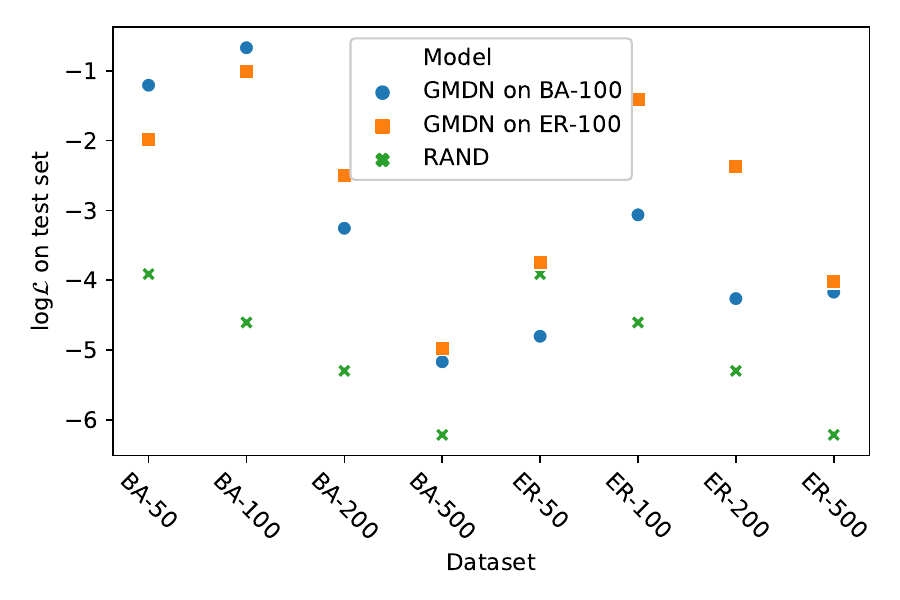}
    \caption{Transfer learning effect of the trained \GMDN{}s are shown as blue dots and orange squares. Higher scores are better. \GMDN{} trained on ER-100 exhibits better transfer on larger BA-datasets, which might be explained by the difficulty of the source task.}
    \label{fig:transfer-effect}
\vskip 0.1in
\end{figure}

\paragraph{Transfer Results}
To tell whether \GMDN{} can transfer knowledge to a random graph of different size and/or family (\ie with different structural properties), we evaluate the trained models on the six additional datasets described in Section \ref{sec:experiments}. Results are shown in Figure \ref{fig:transfer-effect}, where the \RAND{} score acts as the reference baseline. The general trend is that the \GMDN{} trained on ER-100 has better performances than its counterpart trained on BA-100; this is true for all ER datasets, BA-200 and BA-500. This observation suggests that training on ER-100, which we assumed to be a \quotes{harder} task than BA-100 as discussed above, allows the model to better learn the dynamics of SIR and transfer them to completely different graphs. Since the structural properties of the random graphs vary across the datasets, obtaining a transfer effect is therefore not an obvious task.

\begin{table}[!t]
\small
\centering
\begin{tabular}{lcccc}
\toprule
\multicolumn{1}{c}{\multirow{2}{*}{\textbf{Model}}} & \multicolumn{2}{c}{\textbf{alchemy\_full}} & \multicolumn{2}{c}{\textbf{ZINC\_full}} \\
\multicolumn{1}{c}{}                                & $\log \mathcal{L}$     & MAE              & $\log \mathcal{L}$   & MAE             \\ \midrule
RAND                                                & -27.12                 & -                &  -4.20               & -               \\
HIST                                                & -21.91                 & -                &  -1.28               & -               \\
MDN                                                 & -1.36(.90)             & 0.62(.01)        & -1.14(.01)           & 0.67(.00)     \\
DGN                                                 & -7.19(1.3)             & 0.62(.01)        & -0.90(.10)           & 0.49(.03)      \\ \midrule
GMDN                                                & \textbf{-0.57}(1.4)    & \textbf{0.61}(.02)       & \textbf{-0.75}(.10)           & \textbf{0.49}(.04)     \\ \bottomrule
\end{tabular}
\label{tab:results-chemical}
\caption{Results on the chemical tasks show how \GMDN{} consistently reaches better log-likelihood values than the baselines. We also report the MAE as secondary metric for future reference, using the weighted mean of the sub-networks as the prediction (see \citet{bishop_mixture_1994} for alternatives). Clearly, the MAE does not reflect the amount of uncertainty in a model's prediction, whereas the log-likelihood is the natural metric for that matter. Results are averaged over 10 training runs with standard deviation in brackets.}
\end{table}

\begin{figure}[!t]
    \vskip 0.1in
    \centering
    \includegraphics[width=\columnwidth]{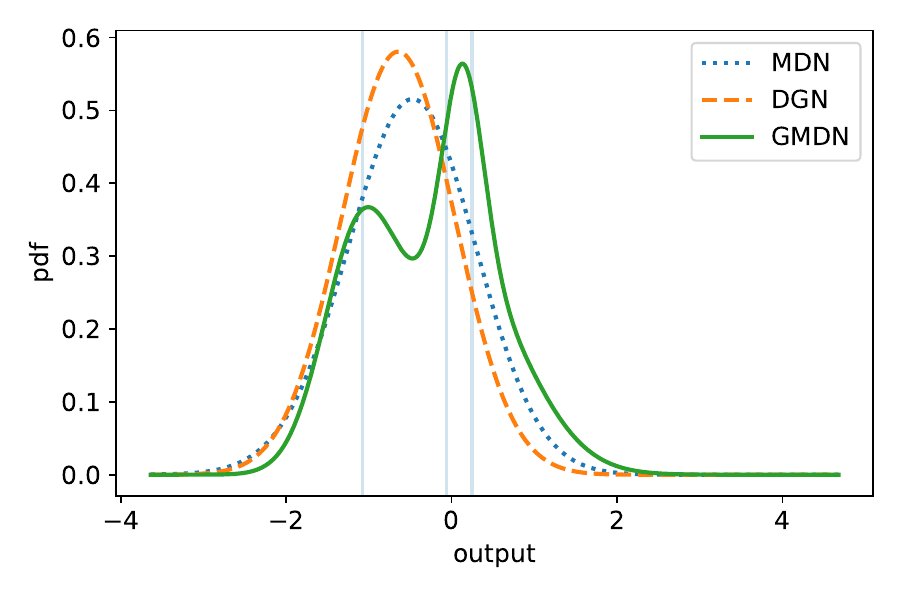}
    \caption{We illustrate the output distributions on the first component, \ie dipole moment, of an alchemy\_full graph. As noted in the text, \DGN{} places high confidence in between the two modes of \GMDN{}. On the contrary, \GMDN{} is able to express uncertainty about the possible output values (vertical lines) associated with isomorphic graphs, which can be found if 3D attributes are not considered. The existence of the two modes suggests that 3D attributes are nonetheless ignored by the three models. See the discussion in Section \ref{sec:experiments} for a more in-depth explanation of the phenomenon.}
    \label{fig:alchemy}
\end{figure}

\paragraph{Chemical Benchmarks}
We conclude this section with results on the real-world chemical benchmarks, which are summarized in Table \ref{tab:results-chemical}. We observe a log-likelihood trend similar to that in Table \ref{tab:results}, with the notable difference that \DGN{} performs much worse than \MDN{} on alchemy\_full. Following the discussion in Section \ref{sec:experiments}, we evaluate how models deal with the uncertainty in the prediction by analyzing one of the output components of alchemy\_full. Figure \ref{fig:alchemy} shows such an example for the first component (dipole moment).
The two modes of the \GMDN{} suggest that, for some input graphs, it may not be clear which output value is more appropriate. This is confirmed by the vertical lines representing output values of isomorphic graphs (as discussed in Section \ref{sec:experiments}). Similarly to Figure \ref{fig:erdos-renyi}, the \DGN{} tries to cover all possible outcomes with a single Gaussian in between the \GMDN{} modes. Although this choice may well minimize the MAE score over the dataset, the \DGN{} fails to model the data we have. In this sense, \GMDN{} can become a useful tool to \textit{(i)} better analyze the data, as uncertainty usually arises from stochasticity, noise, or under-specification of the system of interest, and \textit{(ii)} train deep graph networks which can provide further insights into their predictions and their trustworthiness.

\section{Conclusions}
\label{sec:conclusions}
With the Graph Mixture Density Networks, we have introduced a new family of models that combine the benefits of Deep Graph Networks and Mixture Density Networks. These models can solve challenging tasks where the input is a graph and the conditional output distribution is multimodal. In this respect, we have introduced a novel benchmark application for graph conditional density estimation founded on stochastic epidemiological simulations. The effectiveness of GMDM has also been demonstrated on real-world chemical regression tasks. We believe Graph Mixture Density Networks can play an important role in the approximation of structure-dependent phenomena that exhibit non-trivial conditional output distributions.

\section*{Acknowledgements}
This research was partially supported by TAILOR, a project funded by EU
Horizon 2020 research and innovation programme under GA No 952215. We would like to thank the reviewers for the positive and constructive criticism. We also thank Marco Podda and Francesco Landolfi for their insightful comments and suggestions.

\bibliographystyle{icml2021}
\bibliography{main.bib}

\end{document}